\documentclass[10pt,twocolumn,letterpaper]{article}

\usepackage{cvpr} 
\usepackage{times}
\usepackage{epsfig}
\usepackage{graphicx}
\usepackage{amsmath}
\usepackage{amssymb}
\usepackage{booktabs}
\usepackage{multirow}
\usepackage{url} 

\usepackage[pagebackref=true,breaklinks=true,letterpaper=true,colorlinks,bookmarks=false]{hyperref}

\begin{document}

\title{Trade-offs in Cross-Domain Generalization of Foundation Model Fine-Tuned for Biometric Applications }

\author{Tahar Chettaoui$^{1}$, Naser Damer$^{1,2}$, Fadi Boutros$^{1}$ \\
$^{1}$Fraunhofer Institute for Computer Graphics Research IGD, Germany\\
$^{2}$Technical University of Darmstadt, Germany\\
{\tt\small \{tahar.chettaoui, naser.damer, fadi.boutros\}@igd.fraunhofer.de}
}


\maketitle
\thispagestyle{empty}

\begin{abstract}
Foundation models such as CLIP have demonstrated exceptional zero- and few-shot transfer capabilities across diverse vision tasks. However, when fine-tuned for highly specialized biometric tasks, face recognition (FR), morphing attack detection (MAD), and presentation attack detection (PAD), these models may suffer from over-specialization.  Thus, they may lose one of their foundational strengths, cross-domain generalization. In this work, we systematically quantify these trade-offs by evaluating three instances of CLIP fine-tuned for FR, MAD and PAD.  We evaluate each adapted model as well as the original CLIP baseline on 14 general vision datasets under zero-shot and linear-probe protocols, alongside common FR, MAD and PAD benchmarks. Our results indicate that fine-tuned models suffer from over-specialization, especially when fine-tuned for complex tasks of FR. Also, our results pointed out that task complexity and classification head design, multi-class (FR) vs. binary (MAD and PAD), correlate with the degree of catastrophic forgetting. The FRoundation model with the ViT-L backbone outperforms other approaches on the large scale FR benchmark IJB-C, achieving an improvement of up to 58.52\%. However, it experiences a substantial performance drop on ImageNetV2, reaching only 51.63\% compared to 69.84\% achieved by the baseline CLIP model. Moreover, the larger CLIP architecture consistently preserves more of the model’s original generalization ability than the smaller variant, indicating that increased model capacity may help mitigate over-specialization. 
\url{https://taharchettaoui.github.io/Generalization-of-FM-for-Biometric_web/}
\end{abstract}


\section{Introduction}
Vision and Multi-modal foundation models \cite{DBLP:conf/icml/RadfordKHRGASAM21,DBLP:conf/iccv/CaronTMJMBJ21} have demonstrated remarkable performance across a wide range of computer vision tasks, owing to their ability to learn generalized representations from large-scale, diverse datasets using self-supervised or weakly supervised learning.
Prominent examples such as CLIP \cite{DBLP:conf/icml/RadfordKHRGASAM21},  DINO \cite{caron2021emergingpropertiesselfsupervisedvision}, DINOv2 \cite{oquab2024dinov2learningrobustvisual}, and Segment-anything (SAM)\cite{kirillov2023segment} have demonstrated impressive zero-shot and few-shot generalization capabilities across various visual recognition tasks.

One of the primary benefits of foundation models lies in their ability to transfer knowledge to downstream tasks with minimal task-specific data \cite{DBLP:journals/ivc/ChettaouiDB25}, reducing the reliance on massive training datasets. This property is particularly valuable in domains such as biometrics, where privacy, ethics, or legal challenges may constrain data collection \cite{lirias3838501,DBLP:journals/ivc/BoutrosSFD23} without proper user consent \cite{GDPR_practice}. As a result, foundation models offer a compelling alternative to training specialized models from scratch \cite{DBLP:journals/ivc/ChettaouiDB25,Caldeira_2025_WACV,DBLP:journals/corr/abs-2501-02892,10656894}.
Several recent works have explored the application of foundation models to biometric recognition and its sub-tasks, including template extraction for face verifications \cite{DBLP:journals/ivc/ChettaouiDB25},  attack detections such as morphing \cite{Caldeira_2025_WACV} and presentation attack detections \cite{DBLP:journals/corr/abs-2501-02892}, soft biometric attribute estimation \cite{deandres2024how,hassanpour2024chatgpt},  image quality assessments \cite{10656894}, and biometric decision expandability \cite{deandres2024how}.
However, a critical trade-off emerges: while fine-tuning foundation models enhances performance on the target task, such as presentation attack detection, it may degrade the model's generalizability to other domains or tasks, such as general image classifications, object detection, and image segmentation. This phenomenon, often referred to as catastrophic forgetting or over-specialization \cite{DBLP:journals/corr/abs-2308-08747,DBLP:conf/emnlp/Li0FT24,DBLP:journals/tmlr/MukhotiG0D24}, raises concerns about the robustness and versatility of adapted foundation models. This paper aims at answering the following research question: When foundation models are fine-tuned for a specific downstream task of biometrics, to which degree do these models lose cross-domain and cross-task generalization? To answer this question,  we investigate the trade-offs in cross-domain generalization when CLIP \cite{DBLP:conf/icml/RadfordKHRGASAM21}, a leading multimodal foundation model, is fine-tuned for biometric tasks. We focus on three fine-tuned models:
\begin{itemize}
\setlength{\itemsep}{-0.5em}
    \item FRoundation \cite{DBLP:journals/ivc/ChettaouiDB25}: Fine-tuned CLIP for FR using margin-penalty softmax loss.
    \item MADation \cite{Caldeira_2025_WACV}: Adapted CLIP for morphing attack detection via binary classification.
    \item FoundPAD \cite{DBLP:journals/corr/abs-2501-02892}: Adapt CLIP for presentation attack detection with a task-specific classification head.
\end{itemize}
All these approaches leverage pretrained weights of CLIP and are adapted using Low-Rank Adaptation (LoRA) \cite{hu2021loralowrankadaptationlarge}. We evaluate these models on both biometric benchmarks, including face verification (7  mainstream benchmarks), presentation attack detection, and morphing detection benchmarks (6 main benchmarks), as well as diverse general computer vision datasets (14 testing datasets, e.g., CIFAR-10, ImageNet-v2) to quantify the loss of generalizability in cross-domain settings.
Additionally, we provide the same evaluation results of the base pretrained models, CLIP, without any fine-tuning.
Our analysis includes zero-shot and linear-probe evaluations, revealing how task-specific adaptations impact the model's broader capabilities.
By systematically assessing these trade-offs, this work provides insights into the practical implications of fine-tuning foundation models for biometrics, balancing performance gains with the preservation of generalizability in cross domains/tasks. 
Our results reveal that 1) fine-tuning, as reported in previous works \cite{DBLP:journals/ivc/ChettaouiDB25,Caldeira_2025_WACV,DBLP:journals/corr/abs-2501-02892}, improves the model accuracy on the targeted tasks. However, it loses its cross-domain generalizability. For example, fine-tuning the ViT-L backbone leads to at least a 48.71\% improvement in FR accuracy on the IJB-C benchmark in most cases, but results in an average drop of 5\% on linear-probe evaluation and 13\% on zero-shot evaluation in general vision performance. 2) task complexity (multi-class vs. binary) correlates with the degree of catastrophic forgetting, and 3) larger architectures (ViT-L/14) suffer less generalization loss compared to the smaller architecture (ViT-B/16). These findings contribute to the ongoing research on optimizing foundation models for specialized applications without sacrificing their foundational strengths.


\section{Related Work}
Recent advances in foundation models have raised significant interest in their application to several downstream tasks, including biometrics \cite{DBLP:journals/corr/abs-2108-07258,shahreza2025foundation}.
Chettaoui et al. \cite{DBLP:journals/ivc/ChettaouiDB25} was one of the earliest works utilizing foundation models for FR. 
The work also investigated the performance of foundation models when they are fine-tuned under various data availability scenarios, including using synthetic data to fine-tune foundation models. The work by Chettaoui et al. \cite{DBLP:journals/ivc/ChettaouiDB25} concluded that under a low data availability scenario, fine-tuned foundation models outperformed models trained from scratch for FR. 
MADation \cite{Caldeira_2025_WACV} is a pioneering work that adapts CLIP\cite{DBLP:conf/icml/RadfordKHRGASAM21} for face morphing attack detection. By integrating LoRA\cite{hu2021loralowrankadaptationlarge} layers and a trainable classification head, MADation fine-tunes CLIP's image encoder to detect morphing attacks, outperforming both zero-shot CLIP and models trained from scratch.
FoundPAD \cite{DBLP:journals/corr/abs-2501-02892} also leverages CLIP's pre-trained image encoder with LoRA layers and a binary classifier to detect face presentation attacks. In cross-dataset evaluations, FoundPAD surpasses several state-of-the-art PAD methods.  
Arc2Face \cite{DBLP:conf/eccv/PapantoniouLMDKZ24} utilizes foundation models to extract identity representations from face images and uses them as identity conditions for diffusion models, aiming to generate identity-specific synthetic face images. 
CLIB-FIQA \cite{10656894} leverages the CLIP foundation model to align visual features (from face images) with textual descriptions of quality-related factors (e.g., blur, pose, expression) to assess the quality of face images.

Recent advances in large language models (LLMs), particularly ChatGPT, have prompted an investigation of their performance in the biometric domain. Although originally designed for natural language processing, multimodal extensions of models like GPT-4 have demonstrated competence in visual representation learning tasks. 
Hassanpour et al. \cite{hassanpour2024chatgpt} explore the potential of large language models (LLMs), specifically GPT-4, in performing biometric tasks. The study investigates GPT-4's capabilities in FR, gender detection, and age estimation.
Similarly, DeAndres-Tame et al. \cite{deandres2024how} conducted a comprehensive evaluation of GPT-4 for face verification and soft-biometric attribute estimation. The study focuses on three primary areas: face verification, soft-biometric attribute estimation, and the explainability of decisions made by the model. 
Their study pointed out that GPT-4, despite not being fine-tuned for biometric recognition, achieved up to 94\% verification accuracy on the LFW \cite{huang:inria-00321923} dataset and 96.30\% accuracy for gender classification on the MAAD-Face dataset \cite{DBLP:journals/tifs/TerhorstFKDKK21}.
Kramer \cite{doi:10.1177/03010066241295992} compared GPT-4’s face matching capabilities to human performance using the Glasgow Face Matching benchmarks \cite{burton2010glasgow}. Results indicated that GPT-4 exhibited comparable performance to average human accuracy, demonstrating the model’s suitability for high-level perceptual reasoning despite lacking domain-specific training.
Beyond facial biometrics, Farmanifard and Ross \cite{DBLP:conf/icb/FarmanifardR24} investigated ChatGPT-4's potential in iris recognition tasks. The study employed zero-shot settings to assess the model’s ability for iris recognition.
These previous works \cite{DBLP:journals/ivc/ChettaouiDB25,Caldeira_2025_WACV,DBLP:journals/corr/abs-2501-02892,10656894} demonstrate the potential of foundation models, when properly adapted, to address domain-specific biometric challenges. However, the generalizability of adapted foundation models in cross domains and tasks remains under study, which motivated our work to evaluate cross-domain generalization.

\section{Methodology}
This section introduces the approaches investigated in this work for assessing the loss of generalizability in foundation models when adapted to downstream tasks. First, Section \ref{sec-clip} provides an overview of CLIP, the baseline foundation model used across all considered approaches. Then, Section \ref{sec-clip-ba} describes how CLIP is adapted to different tasks in the field of biometrics. Specifically, we present FRoundation for FR \cite{DBLP:journals/ivc/ChettaouiDB25}, FoundPAD for presentation attack detection (PAD) \cite{DBLP:journals/corr/abs-2501-02892}, and MADation for morphing attack detection (MAD)\cite{Caldeira_2025_WACV}.

\subsection{CLIP} \label{sec-clip}
Contrastive Language Image Pretraining (CLIP), introduced by Radford et al. \cite{DBLP:conf/icml/RadfordKHRGASAM21}, is a multimodal foundation model developed to jointly learn visual and textual representations. It is trained on a large-scale dataset composed of images paired with textual descriptions, enabling the model to align visual and linguistic features through contrastive learning. The architecture includes two independent encoders, one for images and one for text. During training, CLIP optimizes the cosine similarity between embeddings of matching image and text pairs, while reducing the similarity for non-matching pairs. This approach leads to a shared embedding space that supports generalization across a wide range of vision and language tasks \cite{DBLP:conf/icml/RadfordKHRGASAM21}. CLIP employs two distinct architectures for its image encoder: a modified ResNet-50 \cite{DBLP:conf/cvpr/HeZRS16} and the Vision Transformer (ViT) \cite{DBLP:journals/corr/abs-2010-11929}. In this paper, we focus exclusively on the ViT-based variant, as it is adopted as the baseline architecture in the recetnly published biometrics approaches \cite{DBLP:journals/ivc/ChettaouiDB25, DBLP:journals/corr/abs-2501-02892, Caldeira_2025_WACV}. The ViT image encoder largely follows the standard transformer design, with minor modifications such as the addition of a layer normalization to the combined patch and positional embeddings, as well as a slightly altered initialization scheme. The accompanying text encoder is a Transformer \cite{DBLP:conf/nips/VaswaniSPUJGKP17}, which processes tokenized text inputs and projects them into a shared embedding space with visual features. This multimodal architecture enables CLIP to learn strong associations between visual and textual data, supporting its effective transferability to various downstream tasks.

\subsection{Adapting CLIP for Biometric Applications}\label{sec-clip-ba}
The recently published approaches \cite{DBLP:journals/ivc/ChettaouiDB25, DBLP:journals/corr/abs-2501-02892, Caldeira_2025_WACV} considered in this work leverage CLIP \cite{DBLP:conf/icml/RadfordKHRGASAM21}, presented in the previous Section \ref{sec-clip}, and adapt it to their respective downstream tasks using LoRA \cite{hu2021loralowrankadaptationlarge}. LoRA builds on the insight that fine-tuning can be effectively performed in a low-dimensional subspace, as demonstrated in \cite{DBLP:conf/acl/AghajanyanGZ20}, suggesting that pre-trained models lie on a low intrinsic dimension. Rather than updating all weights, LoRA freezes the original parameters and introduces low-rank matrices within each layer. For a weight matrix \( W_0 \in \mathbb{R}^{d \times k} \), the update is defined as \( W = W_0 + BA \), where \( B \in \mathbb{R}^{d \times r} \), \( A \in \mathbb{R}^{r \times k} \), and \( r \ll \min(d, k) \). Only \( A \) and \( B \) are trained, significantly reducing trainable parameters while preserving inference efficiency. Upon completion of fine-tuning, the low-rank adaptation \( BA \) is merged with the frozen base weights \( W_0 \), yielding the final weight matrix \( W \). This allows the model to be deployed without modifications to its inference-time architecture or latency. In this work, we investigate the trade-offs in cross-domain generalization of three approaches that adapt the CLIP foundation model with LoRA to three distinct biometric applications, namely FR, PAD and MAD:

\textbf{FRoundation}
This work \cite{DBLP:journals/ivc/ChettaouiDB25} investigates the application of foundation models to FR tasks, exploring their adaptation across various data availability scenarios, including synthetic data. Results show that while pre-trained foundation models generally underperform compared to models specifically trained for FR, fine-tuning them significantly improves performance. Fine-tuned foundation models often outperform models trained from scratch, especially with limited training data. They fine-tune CLIP, among other foundation models, with LoRA to adapt to the downstream task of FR. To optimize the considered foundation models for FR, they use the widely adopted margin-penalty softmax loss \cite{DBLP:conf/cvpr/DengGXZ19, DBLP:conf/cvpr/WangWZJGZL018}. Specifically, they extend the model architecture with an additional multi-class classification layer and apply CosFace \cite{DBLP:conf/cvpr/WangWZJGZL018} loss as the margin-penalty softmax loss. 

\textbf{FoundPAD} This study \cite{DBLP:journals/corr/abs-2501-02892} adapts the CLIP foundation model for PAD using LoRA. FoundPAD demonstrates strong generalization to unseen domains, achieving competitive results under various data availability scenarios, including with synthetic data. To accomplish this, the architecture is extended with a task-specific classification head, trained using binary cross-entropy loss to differentiate between genuine and PAD images.

\textbf{MADation} This work \cite{Caldeira_2025_WACV} presents an approach that adapts the CLIP foundation model using LoRA to the task of MAD. It was evaluated on multiple MAD datasets and demonstrated competitive performance compared to SOTA methods. To achieve this, the architecture is extended with a task-specific classification head trained using binary cross-entropy loss to distinguish bona fide from morphed facial images. The detection score is obtained from the output of the binary classification layer, with the highest score determining the model’s prediction. All the considered works were made publicly available by the respective authors. Our study uses these official releases as bases for our analyses. 

\section{Experimental Setups} \label{sec-experimental-setup}
In this section, we outline the training datasets and evaluation benchmarks used for the selected models across the considered biometric applications, including FR, MAD, and PAD, along with the additional datasets employed to evaluate the cross-task generalization capabilities of the models.

\subsection{Cross-task Generalization in Non-Biometric Domains} \label{subsec-setup-nonbiometric}

\textbf{Evaluation Datasets} 
To assess the cross-task generalization capabilities of the models, we select a diverse set of benchmarks, each evaluating a different aspect of model performance, following \cite{DBLP:conf/icml/RadfordKHRGASAM21, DBLP:conf/cvpr/ChertiBWWIGSSJ23}. The Rendered SST2 dataset, created by OpenAI \cite{DBLP:conf/icml/RadfordKHRGASAM21}, is designed to measure the optical character recognition (OCR) capability of visual representations. It contains sentences rendered as images with black text on a white background. Country211, also created by OpenAI \cite{DBLP:conf/icml/RadfordKHRGASAM21}, assesses a model’s geolocation prediction abilities. It is derived from the YFCC100M dataset \cite{thomee2016yfcc100m} by selecting images with available GPS coordinates that correspond to country codes. A balanced dataset was constructed by sampling 200 training images and 100 test images for each of the 211 countries. CIFAR-10 and CIFAR-100 \cite{krizhevsky2009learning} are general object classification datasets, each consisting of 60,000 images, with 50,000 used for training and 10,000 for testing. CIFAR-10 includes 10 classes, while CIFAR-100 expands this to 100 classes. These datasets are commonly used to evaluate generalization across diverse object categories. MNIST \cite{lecun1998mnist} is a dataset of 10,000 handwritten digits (0–9) widely used to evaluate basic image classification performance, particularly for digit recognition. Describing Textures in Images (DTD) \cite{DBLP:conf/cvpr/CimpoiMKMV14} is used to evaluate a model’s performance on texture classification. It consists of images from 47 texture categories, with 120 images per category split equally into training, validation, and test sets. SUN397 \cite{DBLP:journals/ijcv/XiaoEHTO16} is a scene recognition dataset used to test generalization across complex scene types with diverse visual characteristics. It contains 108,754 images across 397 categories, representing a wide range of indoor and outdoor environments. FGVC Aircraft \cite{DBLP:journals/corr/MajiRKBV13} is a dataset designed for the classification of aircraft images across 100 different categories. It contains 10,000 images, with 100 images for each of 100 distinct aircraft models. The Stanford Cars dataset \cite{DBLP:conf/iccvw/Krause0DF13} evaluates a model’s ability to discriminate between similar categories with subtle differences. It contains 16,185 images across 196 car classes. Flowers102 \cite{DBLP:conf/icvgip/NilsbackZ08} is an image classification dataset consisting of 102 flower categories, with images that exhibit large-scale, pose, and lighting variations. Food101\cite{DBLP:conf/eccv/BossardGG14} is a data set of 101 food categories with 101,000 images. For each class, 250 test images are provided as well as 750 training images. STL-10 \cite{DBLP:journals/jmlr/CoatesNL11} is an image classification dataset that consists of 10 classes, with 5,000 labeled training images and 8,000 test images. The images were acquired from labeled examples in ImageNet \cite{DBLP:conf/cvpr/DengDSLL009}. The German Traffic Sign Recognition Benchmark (GTSRB) \cite{DBLP:conf/ijcnn/StallkampSSI11} is used to train and evaluate models for traffic sign recognition. It contains 43 distinct traffic sign categories, capturing real-world variations. The dataset contains around 50,000 images in total, including 39,209 training images and 12,630 test images. ImageNet-v2 (INet-v2) \cite{DBLP:conf/icml/RechtRSS19} is a large-scale image classification dataset designed to evaluate the performance of models in real-world scenarios. It contains 10,000 test images distributed across 1,000 classes. The dataset is a re-release of the original ImageNet validation set, but with modifications to improve its robustness by using a more diverse and realistic set of images that better reflect real-world variations, such as changes in lighting, viewpoint, and background. A summary of the datasets used to evaluate various aspects of model generalization, along with their test sizes and number of classes, is presented in Table \ref{tab:datasets-general}.

\textbf{Zero-shot Evaluation} 
CLIP \cite{DBLP:conf/icml/RadfordKHRGASAM21} is pre-trained to predict if an image and a text snippet are paired together in its dataset. To perform zero-shot classification, we reuse this capability and follow the protocol set by the authors \cite{DBLP:conf/icml/RadfordKHRGASAM21}. We use the prompt template “A photo of a \{label\}.” as a default, which helps specify that the text refers to the content of the image, improving performance compared to the baseline of using only the label text. Zero-shot performance can be significantly enhanced by customizing the prompt text for each specific task. For each downstream dataset, we use the shared \cite{DBLP:conf/icml/RadfordKHRGASAM21} set of predefined prompts for each class.  


\textbf{Linear-probe Evaluation} 
For linear-probe evaluation, the CLIP \cite{DBLP:conf/icml/RadfordKHRGASAM21} models adapted with LoRA are used as fixed feature extractors by taking image features from the penultimate layer, excluding any classification layer. To assess performance on downstream tasks, a logistic regression classifier is trained on these fixed features using scikit-learn's L-BFGS optimizer with up to 1{,}000 training iterations, following the setup in \cite{DBLP:conf/icml/RadfordKHRGASAM21}. For linear-probe evaluation, we use the same benchmarks as those used for zero-shot evaluation, with one exception: the ImageNet-v2 \cite{DBLP:conf/icml/RechtRSS19} serves as the test data for the ImageNet benchmark and does not define a separate train split.

\begin{table}[t!]
\begin{center} 
\resizebox{0.88\linewidth}{!}{%
\begin{tabular}{|l|c|c|}
\hline
\textbf{Dataset} & \textbf{Test Size} & \textbf{Number of Classes} \\
\hline
Rendered SST2 & 1,821 & 2 \\
Country211 & 21,100 & 211 \\
CIFAR-10 & 10,000 & 10 \\
CIFAR-100 & 10,000 & 100 \\
MNIST & 10,000 & 10 \\
DTD & 1,880 & 47 \\
SUN397 & 108,754 & 397 \\
FGVC Aircraft & 3,333 & 100 \\
Stanford Cars & 8,041 & 196 \\
Flowers102 & 6,149 & 102 \\
Food101 & 25,250 & 101 \\
STL-10 & 8,000 & 10 \\
GTSRB & 12,630 & 43 \\
ImageNet-v2 & 10,000 & 1,000 \\
\hline
\end{tabular}
}
\end{center}
\caption{Summary of datasets used for evaluating cross-task model generalization, including test size and number of classes.}
\label{tab:datasets-general}
\end{table}

\subsection{Cross-task Generalization in Biometric Domains} \label{subsec-setup-biometric}
This section details the training data, evaluation datasets, and model selection for each approach considered in this study. Table \ref{tab:datasets-training} provides a summary of the training and evaluation datasets used by the considered approaches in \cite{DBLP:journals/ivc/ChettaouiDB25, DBLP:journals/corr/abs-2501-02892, Caldeira_2025_WACV} to train and assess the models for this study.

\textbf{FR: Evaluation Datasets, Model Selection, and Training Data} 
For the FR evaluation, we evaluate the performance of the considered approaches on several widely used FR benchmarks. These include Labeled Faces in the Wild (LFW) \cite{huang:inria-00321923}, Celebrities in Frontal-Profile in the Wild (CFP-FP) \cite{c3517bca662f4193a58fd8f9e3145c8f}, AgeDB30 \cite{moschoglou2017agedb}, Cross-age LFW (CA-LFW) \cite{DBLP:journals/corr/abs-1708-08197}, and CrossPose LFW (CP-LFW) \cite{CPLFWTech}. We report verification accuracies (\%) following the official evaluation protocols for each of these benchmarks. In addition, we evaluated on large-scale evaluation benchmarks, IARPA Janus Benchmark-B (IJB-B) \cite{inproceedingsijbb}, and IARPA Janus Benchmark–C (IJB-C) \cite{DBLP:conf/icb/MazeADKMO0NACG18}. For IJB-C and IJB-B, we used the official 1:1 mixed verification protocol and reported the verification performance as true acceptance rates (TAR) at false acceptance rates (FAR) of $1e-4$ and $1e-5$. These benchmarks were selected because they are commonly used to evaluate the latest advancements in FR and offer a diverse range of use cases \cite{Deng_2022, wang2018cosfacelargemargincosine, ElasticFace, DBLP:conf/iccv/DanLXD0XS23}.
We use the ViT-B/16 and ViT-L/14 architectures as provided by \cite{DBLP:journals/ivc/ChettaouiDB25}. Both are pre-trained CLIP models fine-tuned on the WebFace4M dataset, a subset of the WebFace260M dataset \cite{zhu2021webface260mbenchmarkunveilingpower}, which contains 200K identities and 4 million images. To fine-tune the models, LoRA layers \cite{hu2021loralowrankadaptationlarge} with a rank of 16 are added. 

\textbf{MAD: Evaluation Metrics, Model Selection, and Training Data} 
The MAD evaluation metrics were chosen to ensure alignment with the ISO/IEC 30107-3 \cite{iso30107-3} standard, facilitating consistent benchmarking and comparability with prior studies \cite{DBLP:conf/iwbf/DamerFSKHB23, DBLP:conf/icb/HuberBLRRDNGSCT22}. Performance is reported using three key metrics: the bona fide Presentation Classification Error Rate (BPCER), the Attack Presentation Classification Error Rate (APCER), and the detection Equal Error Rate (EER). The BPCER measures the proportion of genuine images incorrectly classified as attack samples, while the APCER quantifies the proportion of attack images misclassified as genuine. The detection EER represents the error rate at the point where the BPCER and APCER are equal, providing a concise assessment of the system’s overall performance balance. To cover a range of operational points and facilitate comparison, both the APCER at fixed BPCER values and the BPCER at fixed APCER values are reported, evaluated at thresholds of 1\%, 10\%, and 20\%. The benchmarking datasets MAD22 \cite{DBLP:conf/icb/HuberBLRRDNGSCT22} and its extension MorDIFF \cite{DBLP:conf/iwbf/DamerFSKHB23} were selected to ensure the comparability of results and to maintain a domain distinct from the SMDD training data \cite{DBLP:conf/cvpr/DamerLFSPB22}. MAD22 includes morphed face images produced by five different methods: three image-level techniques such as FaceMorpher \cite{quek2019facemorpher}, OpenCV \cite{mallick2016facemorph}, and WebMorph \cite{debruine2018webmorph}, along with two GAN-based, representation-level methods, MIPGAN I and MIPGAN II \cite{DBLP:journals/tbbis/ZhangVRRDB21}. The MorDIFF \cite{DBLP:conf/iwbf/DamerFSKHB23} dataset contains morphing samples generated using a diffusion autoencoder \cite{DBLP:conf/cvpr/PreechakulCWS22}. For each evaluation metric, we report the average score across all six benchmarks.
We select the ViT-B/16 and ViT-L/14 architectures provided by the authors \cite{Caldeira_2025_WACV}. These models are pre-trained CLIP models that have been fine-tuned on the Synthetic Morphing Attack Detection Development (SMDD) dataset \cite{DBLP:conf/cvpr/DamerLFSPB22}, using LoRA\cite{hu2021loralowrankadaptationlarge}. The SMDD dataset is synthetic-based for MAD, consisting of 25,000 bona fide and 15,000 attack images.

\textbf{PAD: Cross-Dataset Evaluation Protocol, Model Selection, and Training Data}
The cross dataset PAD evaluation framework is designed to assess model generalization across diverse real world scenarios by training and testing on different publicly available datasets. This setup includes five well-established PAD benchmarks: MSU MFSD (M) \cite{DBLP:journals/tifs/WenHJ15}, CASIA FASD (C) \cite{DBLP:conf/icb/ZhangYLLYL12}, Replay Attack (I) \cite{DBLP:conf/biosig/ChingovskaAM12}, OULU NPU (O) \cite{DBLP:conf/fgr/BoulkenafetKLFH17}, and CelebA Spoof (CA) \cite{DBLP:conf/eccv/ZhangYLYYSL20}. The evaluation is organized into single source, double source, and triple source scenarios, depending on the number of datasets used during training. In this work, we primarily focus on the triple source cross dataset setting, specifically the O\&C\&M  $\rightarrow$ CA protocol, where models are trained on OULU NPU, CASIA FASD, and MSU MFSD, and evaluated on the large scale and diverse CelebA Spoof dataset. CelebA Spoof\cite{DBLP:conf/eccv/ZhangYLYYSL20} is notably diverse in terms of subjects, illumination, and sensors, and includes four types of attacks: print, replay, three-dimensional mask, and paper cut. The images were collected from the web, resulting in a large-scale dataset with 625,537 images from 10,177 subjects.
For the PAD task, we adopt the ViT-B/16 and ViT-L/14 architectures released by the authors in \cite{DBLP:journals/corr/abs-2501-02892}. These are pre-trained CLIP models further fine-tuned using LoRA\cite{hu2021loralowrankadaptationlarge}, following the O\&C\&M  $\rightarrow$ CA protocol. The OULU-NPU\cite{DBLP:conf/fgr/BoulkenafetKLFH17} dataset is a mobile face PAD dataset comprising 5,940 videos collected from 55 subjects using six different mobile phones. The CASIA-FASD\cite{DBLP:conf/icb/ZhangYLLYL12} dataset contains 600 videos from 50 subjects and features various attack types, including warped photo, cut photo, and video replay attacks. Similarly, the MSU-MFSD\cite{DBLP:journals/tifs/WenHJ15} dataset includes a total of 440 videos from 35 subjects, encompassing both bona fide samples and attack types such as printed photo and video replay. 

\begin{table}
\begin{center}
\resizebox{0.89\linewidth}{!}{%
\begin{tabular}{|l|c|c|}
\hline
\textbf{Approach} & \textbf{Training Data} & \textbf{Test Data} \\ \hline

\multirow{7}{*}{FRoundation\cite{DBLP:journals/ivc/ChettaouiDB25}} &
\multirow{7}{*}{WebFace4M \cite{zhu2021webface260mbenchmarkunveilingpower}} & 
LFW\cite{huang:inria-00321923} \\
&& CFP-FP\cite{c3517bca662f4193a58fd8f9e3145c8f} \\
&& AgeDB-30\cite{moschoglou2017agedb} \\
&& CALFW\cite{DBLP:journals/corr/abs-1708-08197} \\
&& CPLFW\cite{CPLFWTech} \\
&& IJB-B\cite{inproceedingsijbb}  \\
&& IJB-C\cite{DBLP:conf/icb/MazeADKMO0NACG18}  \\ \hline

\multirow{6}{*}{MADation\cite{Caldeira_2025_WACV} } &
\multirow{6}{*}{SMDD\cite{DBLP:conf/cvpr/DamerLFSPB22}} &
MIPGAN I/II\cite{DBLP:journals/tbbis/ZhangVRRDB21} \\ 
&& MorDIFF\cite{DBLP:conf/iwbf/DamerFSKHB23} \\
&& FaceMorpher\cite{quek2019facemorpher} \\
&& OpenCV\cite{mallick2016facemorph} \\
&& WebMorph\cite{debruine2018webmorph} \\ \hline 

\multirow{1}{*}{FoundPAD\cite{DBLP:journals/corr/abs-2501-02892}} &
\multirow{1}{*}{O\&C\&M \cite{DBLP:conf/fgr/BoulkenafetKLFH17, DBLP:conf/icb/ZhangYLLYL12, DBLP:journals/tifs/WenHJ15}} & 
CA \cite{DBLP:conf/eccv/ZhangYLYYSL20} \\       

\hline
\end{tabular}
}
\end{center}
\caption{An overview of the training and evaluation datasets employed by the considered approaches \cite{DBLP:journals/ivc/ChettaouiDB25, DBLP:journals/corr/abs-2501-02892, Caldeira_2025_WACV}.}
\label{tab:datasets-training}
\end{table}

\aboverulesep=0ex
\belowrulesep=0ex
\section{Results} \label{sec-results}
This section presents the results achieved by the considered models across several evaluation benchmarks. Each approach \cite{DBLP:journals/ivc/ChettaouiDB25, DBLP:journals/corr/abs-2501-02892, Caldeira_2025_WACV} delivers two models with distinct architecture, ViT-B and ViT-L, which are pre-trained on CLIP and adapted to their respective tasks using LoRA, as stated in Section \ref{sec-experimental-setup}. The evaluation covers not only general vision benchmarks but also task-specific datasets, including FR, morphing attack detection, and presentation attack detection. All models are evaluated on all considered benchmarks to assess their generalization capabilities across different domains and tasks after being adapted to their respective tasks, in comparison to the original CLIP model. In Section \ref{eval-zs-nbd}, we follow the zero-shot classification approach used by CLIP \cite{DBLP:conf/icml/RadfordKHRGASAM21} to evaluate its cross-domain/task generalization on a wide range of benchmarks. Then, since all considered approaches, FRoundation\cite{DBLP:journals/ivc/ChettaouiDB25}, FoundPAD\cite{DBLP:journals/corr/abs-2501-02892}, and MADation\cite{Caldeira_2025_WACV},  leverage the CLIP image encoder without fine-tuning the text encoder accordingly, we also perform a linear-probe evaluation in Section \ref{eval-lp-generalization}, following \cite{DBLP:conf/icml/RadfordKHRGASAM21}, to assess the ability of the visual representations learned by the image encoder to support cross-domain/task generalization. Finally, in Section \ref{eval-ba}, we present the performance of the original CLIP model, as well as the models adapted to FR, MAD, and PAD, across all considered biometric tasks. We investigate how each model performs on the different benchmarks to assess their generalization across multiple biometric tasks. 


\subsection{Zero-Shot Evaluation} \label{eval-zs-nbd}
We perform zero-shot evaluation across a broad range of benchmarks, defined in Section \ref{subsec-setup-nonbiometric}, to assess the cross-domain/task generalization in non-biometric domains of the approaches under consideration, including the baseline CLIP\cite{DBLP:conf/icml/RadfordKHRGASAM21} foundation model, FRoundation\cite{DBLP:journals/ivc/ChettaouiDB25} adapted for FR, FoundPAD\cite{DBLP:journals/corr/abs-2501-02892} adapted for PAD, and MADation\cite{Caldeira_2025_WACV} adapted for MAD. In addition to the accuracy, we provide a measure of generalization loss after fine-tuning, called the Generalization Loss (GELO), defined as the ratio of the average (AVG) performance after fine-tuning to the average performance before fine-tuning: \( \mathrm{GELO} = \frac{\mathrm{AVG}_{\mathrm{after}}}{\mathrm{AVG}_{\mathrm{before}}} \). Based on the results presented in Table \ref{tab:fr_eval_table}, which shows the accuracy and GELO achieved by each configuration across various evaluation benchmarks, we draw the following observations:

1) When comparing the baseline CLIP to the other approaches adapted with LoRA to their respective biometric tasks, we observe that the baseline CLIP outperforms, as expected, fine-tuned ones on most of the benchmarks, with FRoundation achieving the lowest accuracy among the four approaches. The performance of CLIP is closely followed by MADation and FoundPAD, which exhibit an average performance drop of approximately 1.49\% and a slight improvement of 0.04\%, respectively, across all benchmarks on the ViT-B. Although the CLIP baseline outperforms or closely matches FoundPAD performance on most benchmarks, it shows notably lower performance on three specific datasets using the ViT-B, namely MNIST, Flowers102, and SST2. On these datasets, FoundPAD achieves accuracies of 54.11\%, 71.62\%, and 61.45\%, respectively, while CLIP obtains scores of 51.79\%, 69.96\%, and 60.52\%. It is interesting to note that among these three benchmarks, two, namely MNIST and SST2, are related to OCR tasks. MNIST directly evaluates a model’s ability to perform low-level character and digit recognition \cite{lecun1998mnist, DBLP:conf/icml/RadfordKHRGASAM21}, while SST-2 tests the model’s capacity to leverage OCR for a higher-level semantic task\cite{DBLP:conf/icml/RadfordKHRGASAM21}. The performance can be explained by the fact that CLIP's behavior on OCR tasks has been reported as highly variable\cite{DBLP:conf/icml/RadfordKHRGASAM21}, appearing to be sensitive to both the domain and the type of text being recognized.

2) In a comparison of the adapted approaches to each other using the GELO metric, we observe that FoundPAD outperforms both FRoundation and MADation on average across all benchmarks. FRoundation shows the worst overall performance among the considered approaches, suggesting that the adapted model tends to over-specialize to the FR task relative to other tasks. However, it does not overfit within FR itself, as it achieves better results than all other approaches across a broad range of FR evaluation benchmarks, as shown in Table \ref{tab:eval_table_lp}. FRoundation is trained for FR with a large number of classes, in this case 200K, compared to MADation and FoundPAD, which are trained using binary classification, indicating that optimizing for fine-grained identity discrimination can lead to reduced generalization across tasks.

3) When comparing different model architectures (ViT-L and ViT-B), we observe a clear performance advantage of ViT-L (large model) over ViT-B (relatively small model) across all benchmarks and for all approaches. We observe performance increases of 8.69\%, 14.69\%, 9.78\%, and 8.42\% when using ViT-L instead of ViT-B across all considered benchmarks for CLIP, FRoundation, MADation, and FoundPAD, respectively.

\begin{table*}[!t]
\begin{center}
\resizebox{\linewidth}{!}{%
\renewcommand{\arraystretch}{1.3} 
\begin{tabular}{cc|cccccccccccccccc}
\toprule
\textbf{Approach} & \textbf{Architecture}  
& Food101\cite{DBLP:conf/eccv/BossardGG14} & CIFAR10\cite{krizhevsky2009learning} & CIFAR100\cite{krizhevsky2009learning} & SUN397\cite{DBLP:journals/ijcv/XiaoEHTO16} & Stanford Cars\cite{DBLP:conf/iccvw/Krause0DF13} & FGVC Aircraft\cite{DBLP:journals/corr/MajiRKBV13} & DTD\cite{DBLP:conf/cvpr/CimpoiMKMV14} & MNIST\cite{lecun1998mnist} & Flowers102\cite{DBLP:conf/icvgip/NilsbackZ08} & STL10\cite{DBLP:journals/jmlr/CoatesNL11} & GTSRB\cite{DBLP:conf/ijcnn/StallkampSSI11} & Country211\cite{DBLP:conf/icml/RadfordKHRGASAM21} & SST2\cite{DBLP:conf/icml/RadfordKHRGASAM21} & ImageNet-v2 \cite{DBLP:conf/icml/RechtRSS19} & Avg. & GELO \\

\midrule

\multirow{2}{*}{CLIP\cite{DBLP:conf/icml/RadfordKHRGASAM21}} 
    &  ViT-B/16 & 88.22 & \textbf{90.80} & \textbf{68.22} & \textbf{63.75} & \textbf{64.71} & 24.33 & \textbf{45.64} & 51.79 & 69.96 & 98.25 & 43.33 & 22.22 & 60.52 & \textbf{61.93} & 60.98 & - \\ 
    &  ViT-L/14 & \textbf{93.07} & \textbf{95.57} & \textbf{78.28} & \textbf{67.38} & \textbf{77.94} & \textbf{31.77} &\textbf{ 55.37} & 76.36 & 79.17 & \textbf{99.36} & \textbf{50.55} & \textbf{31.86} & 68.92 & \textbf{69.84} &\textbf{69.67} & - \\  \hline

\multirow{2}{*}{FRoundation\cite{DBLP:journals/ivc/ChettaouiDB25}} 
    & ViT-B/16 & 57.50 & 72.61 & 39.16 & 47.34 & 29.27 & 9.12 & 32.13 & 51.63 & 29.05 & 89.50 & 23.70 & 10.40 & 57.88 & 38.41 & 41.97 &  0.688 \\  
    & ViT-L/14 & 78.92 & 82.49 & 55.41 & 61.97 & 50.35 & 19.23 & 46.49 & 65.95 & 58.86 & 97.10 & 41.53 & 16.16 & 67.11 & 51.63 & 56.66 & 0.813 \\  \hline

\multirow{2}{*}{MADation\cite{Caldeira_2025_WACV}} 
    & ViT-B/16 & 87.30 & 87.41 & 63.81 & 63.00 & 63.15 & 24.21 & 44.26 & 50.82 & 68.66 & \textbf{98.28} & 40.04 & 21.34 & 60.41 & 60.22 & 59.49 &  0.976 \\ 
    & ViT-L/14 & 92.93 & 94.72 & 76.79 & 67.56 & 77.44 & 32.19 & 55.16 & \textbf{78.83} & \textbf{79.31} & 99.35 & 50.32 & 31.40 & 64.20 & 69.59 & 69.27 &  0.994 \\  \hline

\multirow{2}{*}{FoundPAD\cite{DBLP:journals/corr/abs-2501-02892}} 
    &  ViT-B/16 & \textbf{88.31} & 88.83 & 66.15 & 63.65 & 64.30 & \textbf{24.36} & 45.21 & \textbf{54.11} & \textbf{71.62} & 98.17 & \textbf{43.63} & \textbf{22.80} & \textbf{61.45} & 61.66 & \textbf{61.02} & \textbf{1.001}\\ 
    & ViT-L/14 & 93.04 & 95.39 & 78.20 & 67.13 & 77.83 & \textbf{31.77} & 55.21 & 75.58 & 78.83 & 99.33 & 49.61 & 31.83 & \textbf{68.97} & 69.44 & 69.44 & \textbf{0.997} \\ 
\bottomrule
\end{tabular}
}
\end{center}
\caption{Evaluation of zero-shot cross-task generalization across diverse benchmarks, discussed in Section \ref{sec-experimental-setup}, presented as the classification accuracy (\%). CLIP outperforms the fine-tuned models on most benchmarks. FRoundation exhibits the weakest overall performance, indicating that the adapted model tends to over-specialize to the FR task, which limits its ability to generalize to other tasks.}
\label{tab:fr_eval_table}
\end{table*}

\subsection{Linear-Probe Evaluation} \label{eval-lp-generalization}
Since all the considered approaches \cite{DBLP:journals/ivc/ChettaouiDB25, DBLP:journals/corr/abs-2501-02892, Caldeira_2025_WACV} adapt the CLIP \cite{DBLP:conf/icml/RadfordKHRGASAM21} image encoder to their respective biometric task without fine-tuning the corresponding text encoder, we perform a linear-probe evaluation to assess the effectiveness of the visual representations learned by the image encoder in supporting generalization across different domains. Details regarding linear-probe evaluation are outlined in Section \ref{subsec-setup-nonbiometric}. Based on the results in Table \ref{tab:eval_table_lp} that presents the accuracy and GELO, defined in Section \ref{eval-zs-nbd}, achieved by each configuration across several evaluation benchmarks, we made the following observations: 

1) With linear-probe classification, the results achieved by the approaches on the selected benchmarks show a clear improvement over zero-shot evaluation, as expected and reported in \cite{DBLP:conf/icml/RadfordKHRGASAM21}. While zero-shot performance correlates with linear-probe performance, it remains largely sub-optimal. For example, FRoundation achieves 72.61\% on CIFAR-10 and 39.16\% on CIFAR-100 with zero-shot evaluation, and these values increase to 90.79\% and 73.70\%, respectively, when evaluated using linear probing.

2) When comparing the baseline CLIP model to the other approaches adapted with LoRA to their respective biometric tasks, we observe, similar to the zero-shot evaluation (Table \ref{tab:fr_eval_table}), that CLIP outperforms or closely matches all of them on most benchmarks. The gap between CLIP and FoundPAD, both using ViT-B, observed during zero-shot evaluation on the OCM task, specifically for MNIST and SST2, is significantly smaller, even though FoundPAD performs better. 

3) Based on the GELO metric, FoundPAD outperforms both FRoundation and MADation on most benchmarks. A similar trend is observed in zero-shot evaluation, where the over-specialization of FRoundation results in a significantly larger performance gap across different benchmarks compared to FoundPAD. In contrast, MADation closely matches FoundPAD’s performance across benchmarks, indicating better cross-domain generalization. For example, on CIFAR-100 with ViT-B, FRoundation achieves an accuracy of 73.70\%, while FoundPAD reaches 81.23\%. MADation closely matches FoundPAD's performance with 80.46\%, indicating better cross-domain generalization.

4) When comparing different model architectures (ViT-L and ViT-B), similar to zero-shot evaluation, we observe a clear performance advantage of ViT-L over ViT-B across all benchmarks (as was also reported in the CLIP paper \cite{DBLP:conf/icml/RadfordKHRGASAM21}) and for all approaches. The performance gap varies significantly across benchmarks. For example, on CIFAR-10, we observe performance increases of 2.01\%, 2.13\%, 3.07\%, and 2.58\% when using ViT-L instead of ViT-B for CLIP, FRoundation, MADation, and FoundPAD, respectively. In contrast, on FGVC Aircraft, we observe performance increases of 10.29\%, 10.56\%, 11.62\%, and 10.74\% when using ViT-L instead of ViT-B for the same approaches. 


\begin{table*}[!t]
\begin{center}

\resizebox{\linewidth}{!}{%
\renewcommand{\arraystretch}{1.3} 
\begin{tabular}{cc|ccccccccccccccc}
\toprule
\textbf{Approach} & \textbf{Architecture}  
& Food101\cite{DBLP:conf/eccv/BossardGG14} & CIFAR10\cite{krizhevsky2009learning} & CIFAR100\cite{krizhevsky2009learning} & SUN397\cite{DBLP:journals/ijcv/XiaoEHTO16} & Stanford Cars\cite{DBLP:conf/iccvw/Krause0DF13} & FGVC Aircraft\cite{DBLP:journals/corr/MajiRKBV13}& DTD\cite{DBLP:conf/cvpr/CimpoiMKMV14} & MNIST\cite{lecun1998mnist} & Flowers102\cite{DBLP:conf/icvgip/NilsbackZ08} & STL10\cite{DBLP:journals/jmlr/CoatesNL11} & GTSRB\cite{DBLP:conf/ijcnn/StallkampSSI11} & Country211\cite{DBLP:conf/icml/RadfordKHRGASAM21}& SST2\cite{DBLP:conf/icml/RadfordKHRGASAM21} & Avg. & GELO \\

\midrule

\multirow{2}{*}{CLIP\cite{DBLP:conf/icml/RadfordKHRGASAM21}} 
    &  ViT-B/16 & \textbf{92.49} & \textbf{95.99} & \textbf{82.53} & \textbf{82.07} & \textbf{86.33} & \textbf{52.26} & \textbf{75.96} & 98.66 & 95.61 & \textbf{99.17} & 87.64 & 30.58 & 74.19 & \textbf{81.04} & - \\ 
    & ViT-L/14 &\textbf{94.61} & \textbf{98.00} & \textbf{86.71} & 83.69 & 90.72 & 62.55 & \textbf{80.00} & \textbf{99.02} & 98.52 & \textbf{99.78} & 93.25 & 37.15 & \textbf{80.50} & \textbf{84.96} & - \\  \hline

\multirow{2}{*}{FRoundation\cite{DBLP:journals/ivc/ChettaouiDB25}} 
    & ViT-B/16 & 83.19 & 90.79 & 73.70 & 75.93 & 78.31 & 46.23 & 70.00 & 98.38 & 89.48 & 95.84 & 86.99 & 19.80 & 70.01 & 75.28 & 0.929 \\ 
    & ViT-L/14 & 90.27 & 92.92 & 76.16 & 80.87 & 89.97 & 56.79 & 74.79 & 98.84 & 97.09 & 99.05 & 91.85 & 23.15 & 78.31 & 80.77 & 0.951 \\  \hline

\multirow{2}{*}{MADation\cite{Caldeira_2025_WACV}} 
    &  ViT-B/16 & 92.06 & 94.68 & 80.46 & 81.73 & 85.40 & 51.90 & 75.79 & 98.68 & 95.88 & 99.06 & 84.19 & 30.01 & 74.14 & 80.31 & 0.991 \\ 
    & ViT-L/14 & 94.38 & 97.75 & 86.20 & \textbf{83.88} & \textbf{90.93} & \textbf{63.52} & \textbf{80.00} & 98.96 & 98.42 & 99.76 & 93.19 & 36.72 & 80.06 & 84.91 & \textbf{0.999} \\  \hline

\multirow{2}{*}{FoundPAD\cite{DBLP:journals/corr/abs-2501-02892}} 
    &  ViT-B/16 & 92.27 & 95.31 & 81.23 & 81.99 & 86.03 & 51.78 & 75.43 & \textbf{98.69} & \textbf{95.69} & 99.02 & \textbf{88.08} & \textbf{30.63} & \textbf{74.24} & 80.80 & \textbf{0.997}  \\ 
    & ViT-L/14 & 94.48 & 97.89 & 86.55 & 83.79 & 90.72 & 62.52 & 79.73 & 98.97 & \textbf{98.55} & \textbf{99.78} & \textbf{93.29} & \textbf{37.17} & 80.34 & 84.91 & \textbf{0.999} \\ 
\bottomrule
\end{tabular}
}   
\end{center}
\caption{Evaluation of linear-probe cross-task generalization across diverse benchmarks, discussed in Section \ref{sec-experimental-setup}, presented as the classification accuracy (\%). While zero-shot performance shows a correlation with linear-probe accuracy, it remains largely sub-optimal.}
\label{tab:eval_table_lp}
\end{table*}

\subsection{CLIP Performance on Biometric Applications} \label{eval-ba}

\begin{table*}[!t]
\begin{center}
\resizebox{\linewidth}{!}{%
\renewcommand{\arraystretch}{1.3} 
\begin{tabular}{cc|
c c c c c c c|
cc| 
c ccc ccc}
\toprule
\multirow{3}{*}{\textbf{Approach}} & \multirow{3}{*}{\textbf{Architecture}} 
& \multicolumn{7}{c|}{\textbf{FR Benchmarks}} 
& \multicolumn{2}{c|}{\textbf{PAD Benchmarks}} 
& \multicolumn{7}{c}{\textbf{MAD Benchmark}} \\
\cmidrule(lr){3-9} 
\cmidrule(lr){10-11} 
\cmidrule(lr){12-18} 
& & \multirow{2}{*}{LFW\cite{huang:inria-00321923}} & \multirow{2}{*}{CFP-FP\cite{c3517bca662f4193a58fd8f9e3145c8f}} & \multirow{2}{*}{AgeDB-30\cite{moschoglou2017agedb}} 
& \multirow{2}{*}{CALFW\cite{DBLP:journals/corr/abs-1708-08197}} & \multirow{2}{*}{CPLFW\cite{CPLFWTech}} & \multirow{2}{*}{IJB-B\cite{inproceedingsijbb}} & \multirow{2}{*}{IJB-C\cite{DBLP:conf/icb/MazeADKMO0NACG18} } 
& \multicolumn{2}{c|}{CA \cite{DBLP:conf/eccv/ZhangYLYYSL20}} 
& \multirow{2}{*}{EER} 
& \multicolumn{3}{c}{APCER @ BPCER} 
& \multicolumn{3}{c}{BPCER @ APCER} \\
\cmidrule(lr){10-11} 
\cmidrule(lr){13-15} 
\cmidrule(lr){16-18} 
& & & & & & & 
& & HTER $\downarrow$ & AUC $\uparrow$
& & 1.00 & 10.00 & 20.00 
& 1.00 & 10.00 & 20.00 \\
\midrule
\multirow{2}{*}{CLIP\cite{DBLP:conf/icml/RadfordKHRGASAM21}} 
    &  ViT-B/16 &93.33 & 88.86 & 74.67 & 77.13 & 79.23 & 27.79 & 32.40 & 43.95 & 58.36 & 23.59 & 94.01 & 48.67 & 30.24 & 65.44 & 38.97 & 26.80 \\ 
    & ViT-L/14 & 95.90 & 90.66 & 79.82 & 83.10 & 82.73 & 40.90 & 44.69 & 31.37 & 74.83 & 12.13 & 43.43 & 18.50 & 9.35 & 43.87 & 16.01 & 8.99 \\   \hline

\multirow{2}{*}{FRoundation\cite{DBLP:journals/ivc/ChettaouiDB25}} 
    & ViT-B/16 & 99.30 & 93.93 & 88.90 & 92.75 & 90.67 & 81.52 & 85.63 & 49.62 & 50.47 & 51.56 & 99.98 & 93.80 & 85.84 & 99.10 & 91.58 & 84.64\\  
    & ViT-L/14 & 99.65 & 96.50 & 93.72 & 94.37 &93.73 & 90.72 & 93.40 & 49.29 & 51.23 & 49.71 & 99.88 & 90.28 & 81.95 & 99.43 & 90.03 & 79.74 \\ \hline

\multirow{2}{*}{MADation\cite{Caldeira_2025_WACV}} 
    &  ViT-B/16 & 86.48 & 83.36 & 64.38 & 68.78 & 72.53 & 15.99 & 19.70 & 46.73 & 54.35 & 11.89 & 39.36 & 17.18 & 12.39 & 42.64 & 22.14 & 13.97 \\ 
    & ViT-L/14 & 89.07 & 88.26 & 74.93 & 74.92 & 77.13 & 30.54 & 34.88 & 39.08 & 65.90 & 11.94 & 29.24 & 14.05 & 11.04 & 60.40 & 18.90 & 7.57 \\  \hline

\multirow{2}{*}{FoundPAD\cite{DBLP:journals/corr/abs-2501-02892}} 
    &  ViT-B/16 & 90.57 & 87.27 & 72.28 & 74.50 & 76.13 & 25.54 & 30.69 & 15.6 & 91.0 & 30.35 & 98.32 & 76.98 & 48.30 & 73.61 & 49.84 & 38.73\\ 
    &  ViT-L/14 & 95.73 & 88.18 & 78.60 & 81.97 & 82.90 & 41.13 & 44.51 & 43.0 & 59.7 & 32.32 & 94.63 & 68.82 & 49.09 & 73.04 & 50.49 & 41.58 \\ 

\bottomrule
\end{tabular}
}
\end{center}
\caption{Baseline and dedicated biometric models performance on FR, PAD, and MAD benchmarks, discussed in Section \ref{sec-experimental-setup}, all metrics presented in (\%). Each approach adapted to a specific biometric task outperforms the others within its domain. While FRoundation achieves better results on FR benchmarks, it underperforms in generalization over other biometric tasks, revealing the limitations of task-specific over-specialization. Similar conclusions can be made for MADation and FoundPAD.}
\label{tab:table_biometrics}

\end{table*}

In this section, we systematically evaluate the cross-task generalization of the considered approaches within the biometric domains. This includes the baseline CLIP, FRoundation adapted for FR, FoundPAD adapted for PAD, and MADation adapted for MAD. The evaluation is carried out across a wide range of benchmarks for each domain, as detailed in Section \ref{sec-experimental-setup}. From the results shown in Table \ref{tab:table_biometrics}, which outlines the accuracy of each configuration across various FR, MAD, and PAD evaluation benchmarks, we observed the following:


1) As expected, each approach adapted to a specific biometric task outperforms the others within its domain. Notably, the baseline CLIP model consistently ranks second, demonstrating strong cross-domain generalization capabilities. For FR, the baseline CLIP model achieves 44.69\% on IJB-C as TAR at a FAR of $1e-4$ using the ViT-L backbone. While this is notably lower than FRoundation, which achieves 93.40\% when fine-tuned on WebFace4M using LoRA, CLIP still outperforms FoundPAD with 44.51\% and MADation with 34.88\%, demonstrating its strong generalization despite not being optimized for the task. A comparable analysis can be conducted for both MAD and PAD, revealing consistent patterns in performance and task-specific effectiveness.


2) When evaluating the considered approaches on the FR task, a significant performance gap is observed across both large-scale benchmarks such as IJB-B and IJB-C, and smaller-scale datasets including LFW, CFP-FP, AgeDB-30, CALFW, and CPLFW. For example, on the IJB-C benchmark, the FRoundation model with the ViT-L backbone consistently outperforms the other approaches by at least 48.71\%, achieving scores up to 58.52\%. It is important to highlight that in the cross-task generalization experiments conducted on non-biometric domains, using both zero-shot and linear-probe evaluation settings, FRoundation consistently performs worse than the other approaches, as presented in Section \ref{eval-zs-nbd}. This is also the case for biometric tasks, where FRoundation consistently produces the weakest results compared to the other approaches in this study, on the MAD and PAD evaluation benchmarks. This shows the complexity of the FR task and indicates that achieving near SOTA performance requires a high level of task-specific over-specialization, often limiting the model's ability to generalize across diverse domains.

3) In contrast to the performance disparities observed in FR, the performance gap in MAD is less pronounced. While FRoundation and FoundPAD show results that fall significantly short of the performance achieved by MADation, the baseline CLIP model closely approaches the results of MADation. It is crucial to note that MADation was specifically trained for the MAD downstream task, incorporating an additional trained binary classification head to distinguish between bona fide and morphed facial images. In contrast, the MAD evaluation for the other approaches was conducted in a zero-shot setting. This is observed for the ViT-L backbone, where the baseline CLIP achieves an EER of 12.13\%, while MADation reaches 11.94\%. In comparison, the other approaches perform significantly worse, with FoundPAD and FRoundation achieving EERs of 32.32\% and 49.71\%, respectively.



\section{Conclusion}
This paper provides the first comprehensive quantification of over-specialization in the CLIP foundation model when adapted to biometric domains. We evaluated three instances of CLIP fine-tuned for FR (FRoundation), MAD (MADation), and PAD (FoundPAD) on their corresponding target benchmarks and general vision benchmarks. We assessed the over-specialization of the fine-tuned models by comparing their performances with baseline CLIP models. Our results showed that fine-tuned models achieved significant gains on their target biometric benchmarks while exhibiting a loss in general vision tasks. 
For example, FRoundation achieved 93.40\% TAR@FAR on the IJB-C benchmark compared to 44.69\% achieved by CLIP. However,  CLIP achieved 69.84\% accuracy on ImageNetV2, in comparison to 51.63\% achieved by FRoundation.
Additionally, the larger architecture of CLIP consistently retains more of CLIP’s baseline generalization compared to the smaller architecture, suggesting that capacity can buffer against over-specialization. These insights highlight the need for future foundation model adaptations that strike an optimal balance between task-specific accuracy and broad generalization.

\section*{Acknowledgments}

This research work has been funded by the German Federal Ministry of Education and Research and the Hessen State Ministry for Higher Education, Research and the Arts within their joint support of the National Research Center for Applied Cybersecurity ATHENE.

{\small
\bibliographystyle{ieee}
\bibliography{egbib}
}

\end{document}